# *No comments*
## Addressing commentary sections in websites' analyses

Florian Cafiero*1,4, Paul Guille-Escuret1,2*†, Jeremy K. Ward2,3

1- GEMASS (CNRS, Sorbonne Université), Paris, France
2- Vitrome (Aix Marseille Université, IRD, AP-HM, SSA) Marseille, France
3- CERMES3 (INSERM, CNRS, EHESS, Université de Paris), Villejuif, France
4- Medialab (Sciences Po Paris), Paris, France
*These authors contributed equally.
† Corresponding author: paulguilleescuret@gmail.com

**Abstract**

Removing or extracting the commentary sections from a series of websites is a tedious task, as no standard way to code them is widely adopted. This operation is thus very rarely performed. In this paper, we show that these commentary sections can induce significant biases in the analyses, especially in the case of controversial topics. We draw on a corpus of vaccine critical websites to show why it is key to eliminate such commentary sections – or to keep them for a separate analysis. We show that the commentary sections of vaccine critical websites are often raided by opponents and embarrassing allies, whose contributions should not be treated as if they came from the author(s) of the page. We then propose a semi-automated method to remove or extract commentary sections.

**Highlights**

- Commentary sections can induce biases in the analysis of websites' contents
- Analyzing these sections can be interesting *per se*.
- We illustrate these points using a corpus of anti-vaccine websites.
- We provide guidelines to remove or extract these sections.

Studying corpora of websites, through methods such as topic modeling, sentiment analysis or hyperlink analysis, is increasingly adopted in the social sciences (Froio, 2018 ; Marres, 2015), information science (Bounegru et al., 2017 ; Bennett et al. 2011), and digital humanities (Severo et al., 2018; Romele et al. 2016 ; Berthelot et al. 2016). Whatever the data collection strategy adopted, web crawling systematically induces some *noise*, i.e., some "unnecessary, irrelevant information in a data set" (Waldherr et al., 2017, p.2). In most cases, filtering out this noise means selecting some issue-specific contents, a difficult task, but for which various strategies are now available, such as keyword filtering or machine learning (Waldherr et al., 2017). Besides these topic-related issues, other factors can induce systematic noise. The presence of contents produced by individuals other than the owner of the page, notably through the "commentary sections", is one of them.

Several papers have focused on the contents of commentary sections opened by online media (Wu & Atkin, 2018; Toepfl & Piwoni, 2015), comments left by users on social media (Faase et al., 2016; Harlow, 2012), or on comparing these two types of contents (Su et al., 2018). Some studies have examined comment sections posted on a single website (Pihlaja, 2016). But to our knowledge, the treatment and analysis of commentary sections from a corpus of different websites is still an area to explore.

This can be explained by the serious technical difficulties raised by extracting commentary sections from a diversity of websites. Many languages can be used to encode pages: HTML 4.0, HTML 5.0, XHTML, Ajax, Ruby on Rails etc. Some standards obviously exist, for instance for some blog platforms, but they are not widely adopted. Unexpected means to open a commentary section, such as considering commentary sections as a subpart of a forum, can occur as well.

A simple solution would be to focus on extracting only the easily retrievable commentary sections. However, most of the time, this shortcut should not be an option. In the context of social analyses, filtering content according to production techniques would almost always induce important biases. The way the commentary section is encoded is in itself a socially meaningful phenomenon: it demonstrates the user's literacy in web programming, or her financial means. Excluding very poorly encoded pages, or virtuoso contents written by expert programmers, could thus mean excluding specific social groups from further analysis.

In this paper, we advocate the importance of addressing these commentary sections, which can considerably bias analyses, but can also be interesting to study for themselves. To illustrate our point, we analyze a set of French-speaking vaccine-critical websites, from which we extracted the commentary sections. We then give insights on our removal/extraction method, and provide its current implementation.

## A noise-inducing section

When analyzing web pages, leaving commentary sections implies accepting a certain level of noise. These parts of the websites add links and texts that have not been produced - and in many cases that would not be endorsed - by the owner of the site.

But is this noise always likely to significantly bias analyses ? Our latter remark seems at least especially relevant when studying controversial topics, one where opposing camps form and are likely to go on each other's websites.

**Our case study: French-speaking vaccine-critical activists**

As part of a project on the relationship between vaccine-critical actors, we constituted a dataset of 254 French-speaking websites (27292 HTML pages) of individuals, associations or political parties, who criticized at least one vaccine officially recommended by French public health authorities (Cafiero et al., 2021). Collected in March 2017, this database contains contents produced along the past decade.

After reading its content (excluding commentaries), each website was annotated depending on:
- the importance given to the theme of vaccination: is the website dedicated to vaccines (single-issue website), or are vaccines one of many themes broached on the website (multi-issue website)?
- the radicality of the criticism: does the website criticize vaccination in general (the principle of vaccination in itself, or more than 10 vaccines: radical critique) or does it criticize only some vaccines (reformist critique)?

The websites were thus divided into four categories:

|  | Radical | Reformist |
|---|---|---|
| Single Issue | Single-Issue Radical (SIRA) | Single-Issue Reformist (SIRE) |
| Multi Issue | Multi-Issue Radical (MIRA) | Multi-Issue Reformist (MIRE) |

*Table 1 - Typology of vaccine-critical actors applied to our corpus of websites. Coloring is consistent with the following graphs and tables.*

One of the main purposes of our study was to understand the relationships between these categories of actors. The most common rationale for the current surge in vaccine hesitancy is that Internet use would have helped radical critiques (science deniers, conspirationists…), actually active in France (Gargiulo et al., 2020), reach an ever-growing audience. This explanation is however challenged by the growing success of critics who present themselves as being different from « anti-vaxxers », focusing on an apparently less radical agenda, coining slogans such as « green our vaccines ». Do these new actors and the more radical vaccine critical actors form a cohesive community ? Or is there a fragmentation between these various groups ? The « multi-issue reformists » were of particular interest to us, as they group together many major actors of the vaccine controversies in France, at least in terms of visibility. Understanding their relationship with other types of vaccine-critical activists was thus key to our study.

**Hyperlink analysis**

In our example, when leaving the commentary sections, mutual hyperlinks between our websites yield the network displayed on the left of figure 1. When looking at this network, it would seem that the various vaccine-critical groups are indeed tightly interconnected, no matter the diversity of their speeches or opinions.

Six multi-issue reformists (in pink) are separated from the main graph, but most of them are connected to it. Yet, when removing the commentary sections, the outcome significantly changes: now, a majority of multi-issue reformists are not connected to the main graph - which should lead the analyst toward a very different conclusion.

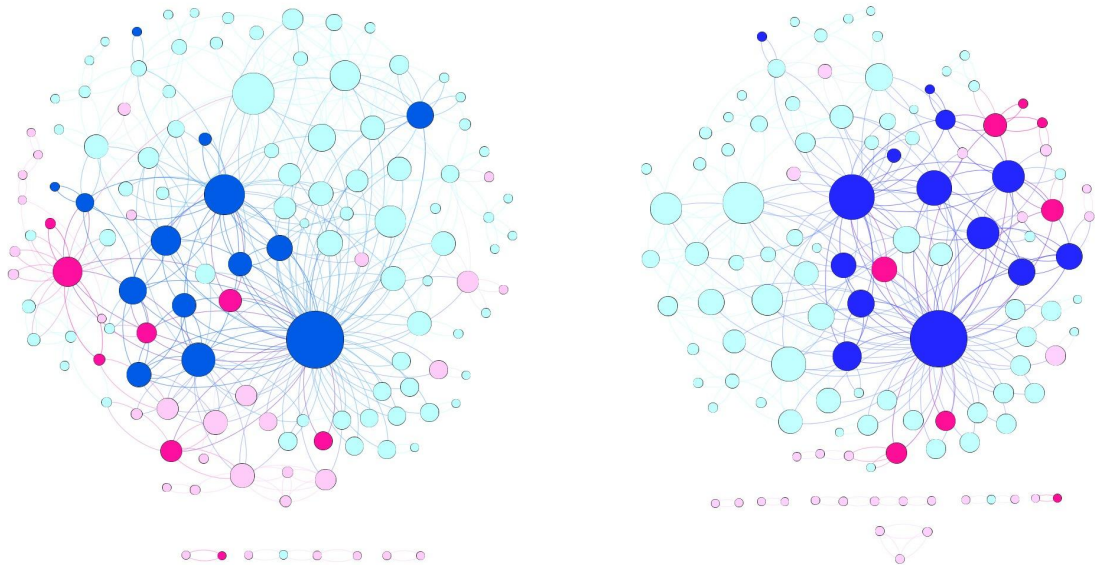

*Fig 1 - Mutual links between French-speaking vaccine-critical websites, with commentary section (left) and without (right). In blue, radical anti-vaccine, in red reformist anti-vaccine. Spatialization through Force Atlas 2 algorithm (Jacomy et al., 2014) with Gephi (Bastian et al., 2009). The size of a node is proportional to the number of appearances in the media.*

Hyperlinks posted in our websites' commentary sections are quite numerous. More importantly, they do not generate an evenly-spread white noise. The hyperlinks appearing in the comments are indeed very unevenly distributed across our various categories of actors (table 2).

| Edge type | Outside commentary sections | Inside commentary sections | Percentage of links from the commentary sections |
| --- | --- | --- | --- |
| MIRE → SIRA | 29 | 19 | 0,40 |
| MIRE → MIRA | 40 | 25 | 0,38 |
| MIRE → SIRE | 24 | 7 | 0,23 |
| MIRE → MIRE | 101 | 23 | 0,19 |
| SIRE → MIRA | 20 | 4 | 0,17 |
| SIRE → MIRE | 36 | 4 | 0,10 |
| MIRA → MIRA | 687 | 44 | 0,06 |
| MIRA → MIRE | 276 | 17 | 0,06 |
| SIRA → MIRA | 118 | 7 | 0,06 |
| MIRA → SIRA | 278 | 15 | 0,05 |
| SIRE → SIRA | 21 | 1 | 0,05 |
| MIRA → SIRE | 102 | 3 | 0,03 |
| SIRA → SIRA | 81 | 0 | 0,00 |
| SIRA → SIRE | 39 | 0 | 0,00 |
| SIRA → MIRE | 91 | 0 | 0,00 |
| SIRE → SIRE | 18 | 0 | 0,00 |

*Table 2 - Links between French-speaking vaccine-critical websites, outside commentary section and inside commentary section. The last column indicates the proportion of hyperlinks found in the commentary section in the total number of hyperlinks between the types of actors indicated.*

Reformists, and especially multi-issue reformists, seem to receive many comments including hyperlinks pointing to other websites of our database, while it is almost never the case for radicals' websites. The magnitude of this phenomenon is far from negligible. Circa 40% of the links from Multi Issue Reformists to Radical websites are indeed located in commentary sections. On the other hand, links from Single Issue Radicals to Reformists are always in the main page, and never in the commentary section.

Any visual inspection or statistical analysis including the commentary sections would thus overestimate the cohesion of the landscape of vaccine criticism. In particular, it would give the impression that reformists are much more connected to the rest of the actors than they actually are.

**Text mining**

The damage caused to text analysis by the commentary sections can sometimes be obvious, and some aspects are likely to be eliminated by standard post-treatments. It is the case for instance when spam filters are not activated, or malfunction. Lexically speaking, a website dedicated to aluminum as an adjuvant for vaccines (and its purported risks) can look like an online textile shop if comments are included in the analysis as in the example described in fig. 2.

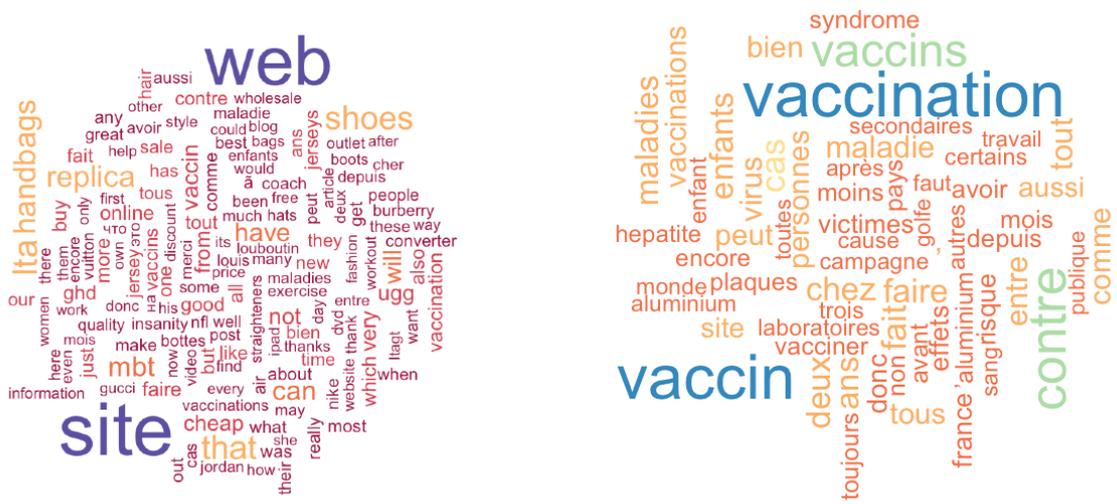

*Fig 2 - On the left, word cloud of the most frequent tokens on a website campaigning against the use of aluminum in vaccines (French stopwords and numbers removed, using the R package TM (Feinerer, 2013)), without removing the comments ; on the right, result of the same treatment applied to the same website, when removing the*

*commentary sections following our method. The size of a word is proportional to its frequency.*

For this particular website, further treatment could have led to the elimination of the contents of the spam. Explicit content filtering for topic relevance, through lexical field queries or machine learning (Waldherr et al., 2017), would probably have been efficient, depending on the specific implementation chosen. Yet, even these precautions could prove to be useless when the topics studied are somehow congruent with the contents of the spam. Had we studied fashion websites, our filters would have kept most of the spam contents found on this website. And had the spam advertised for drugs (which happens frequently) instead of clothes, our filters could have considered them as relevant to our study on health and vaccines. We could thus have ended up commenting on how some specific vaccine-critical activists also seem involved in the controversy surrounding painkillers, as it could have been a topic emerging on some websites.

Beyond spams, the contents of the commentary section can lead to biases that are harder to detect. Most comments are likely to actually deal with the same broad topics as the page they are posted on. Content filtering would thus fail to eliminate them. This problem could seem harmless: if the comments deal with the same matters as the content of the page, they are unlikely to cloud our results, for instance when we run a topic modeling algorithm. Unfortunately, even this statement proves to be wrong. When examining the main topics detected in a prominent vaccine-critical blog (table 3), major differences emerge between the version with comment sections and the one without them. Both computations detect that this blog talks about the vaccine against measles ("*rougeole*" in French) - topic 3 in both tables -, and criticizes vaccination against influenza (*"grippe"* in French) - topic 2 with commentary sections, topic 4 without commentary sections. Analysis of the cleaned database shows a specific interest in controversies surrounding the 2009 H1N1 influenza vaccine (*pandemrix*) (topic 5). The debate about the role of medical doctors in administering vaccines, and the new policy enforced on that matter by Roselyne Bachelot, then minister for health, (topic 1) also seem to be of considerable interest to this blogger - who happens to be a medical doctor. These two topics disappear when the commentary sections are left in the database. In their place, two controversies about other vaccines appear. Topic 4 deals with the vaccine against hepatitis B, and the age at which it is administered. In the same vein, topic 5 deals with the human papillomavirus vaccine (*gardasil*) and the debate about vaccinating against a sexually transmissible disease at a young age. These questions are actually addressed by the website's owner, but are not as important as the other topics. Yet, these last two topics have been very widely commented, and thus now appear as major themes of the website. They obliterate topics that are actually important to the author, but less discussed by his readers.

**With commentary sections**

|   | Topic 1 | Topic 2 | Topic 3 | Topic 4 | Topic 5 |
|---|---|---|---|---|---|
| 1 | dit | grippe | cas | vaccin | vaccin |
| 2 | fait | vaccination | vaccination | ans | gardasil |
| 3 | tout | décès | vaccin | contre | cancer |
| 4 | non | experts | pays | vaccins | vaccination |
| 5 | être | personnes | contre | vaccination | jeunes |
| 6 | faire | risque | rougeole | hépatite | ans |

**Without commentary sections**

|   | Topic 1 | Topic 2 | Topic 3 | Topic 4 | Topic 5 |
|---|---|---|---|---|---|
| 1 | médecins | vaccin | cas | grippe | vaccins |
| 2 | faire | vaccins | vaccination | vaccination | santé |
| 3 | bachelot | effets | vaccin | chez | tous |
| 4 | madame | comme | rougeole | non | false |
| 5 | tout | contre | enfants | personnes | pandemrix |
| 6 | quand | indésirables | pays | efficacité | intérêt |

*Table 3 - Results of a Latent Dirichlet Allocation topic modeling (Blei et al., 2003) with Gibbs Sampling, led on the pages of "docteur du 16" 's blog. At the top, results without removing the commentary sections. At the bottom, results after having removed the commentary sections.*

Without removing the commentary sections, we would have concluded that our vaccine critical actors form a tightly-knit community, very interested in controversies around HPV and hepatitis B vaccines, but who ignore major issues such as the H1N1 vaccination campaign. We probably would have filtered out some pages riddled with spam contents, and missed their texts regarding aluminum in vaccines - a major topic of vaccine controversies in France. All this goes far beyond acceptable noise, and suggests the importance of implementing data wrangling practices.

## A method for deleting or extracting comments

We provide here the method we implemented to extract the commentary sections of our very diverse corpus of websites. Our method entails working on the full code of the pages in the corpus The associated code is available at: [https://github.com/floriancafiero/nocomments](https://github.com/floriancafiero/nocomments).

Comments can be considered as a section as a whole, or as singular entities, depending on the choice of hypotheses and perspective. We thus propose two options: rough slicing (isolating the entire comment section for extraction or removal) and precise slicing (isolating each comment and its metadata, for further analysis). In the latter case, it is not necessary to start by performing rough slicing then move on to precise slicing. Yet, the safest strategy might often be to do so, and thus have a better grasp on the sources of possible errors.

In both cases, we must first inspect the .html code behind the page, find the relevant sequence of code delimiting the content we are looking for, and transcribe them in an *encoding file*. From then on, our code perfoms the deletion or extraction of the comments. We provide on our github repository a series of exemples of delimiters we encountered, including those present on some common websites (blogspot, over blog, canalblog etc.).

**Rough slicing**

This function targets full comment sections, and helps delete or extract them directly from the *html,* as well as the hyperlinks they include.

The main issue is to find a unique pattern corresponding to the opening of comment sections and one, not necessarily unique, corresponding to its end. In most cases it's possible to use the package rvest (Wickham, 2016), navigator extensions like SelectorGadget, or one of your own favorite packages to identify tags associated with nodes corresponding to comment sections. But as some blogs and websites belong to the realm of craftsmanship more than standardized production, direct reading of html files to understand comment section delimitations might be necessary. Moreover, to test the extraction process, it is obviously necessary to find a page with at least one comment which might be difficult for some low traffic websites. In addition to site ids (1) and labels (2), the encoding file we used include :

- a logical variable indicating the presence of comment sections (3)
- opening patterns (4)
- closing patterns (5).

*Precise slicing*

The function directly targets each comment, extracting separately its content and main metadata.

In addition to website ids (1) and labels (2), encoding file then includes :

- an "empty size" variable (3) corresponding to the size of the code fragment for a comment section without comments,
- full comment extraction patterns (4),
- date of the comment extraction patterns (5)
- name of the commenter extraction patterns (6)
- depth of the comment extraction patterns (6) indicating if a comment is a response to previous ones
- website of the commenter extraction patterns (7)
- text of the comment extraction patterns (8)

If one element is missing, the corresponding variable is coded as "False".

## Quality of the results

The quality of the results through this process of course depends on the quality of the crawl. A few standard flaws in web crawler can lead to unfortunate errors. For example, some comment sections are not fully displayed on the webpage - as an "unroll" option redirect to sub-pages - and the crawler is not able to unroll.

Some comment sections may rely on Facebook contents, or on Java parts, which are generally not properly harvested. In those cases, hyperlinks and text of missing comments will not contaminate the database, but their content cannot be analyzed.

Beyond these limitations, users can perform a few operations to help minimize errors. As comment extraction may sometimes fail, it is essential to go back and forth between extracting comments, and testing if the delimiters we isolated helped us get the results we expect. If all extracted comment sections have the same size, either the extraction failed or the website does not have a single comment. The [extract_comment_sections](extract_comment_sections) function automatically produces an error report to help the user spot the most standard problems: absence of comment section opening, absence of comment section closure and multiple openings.

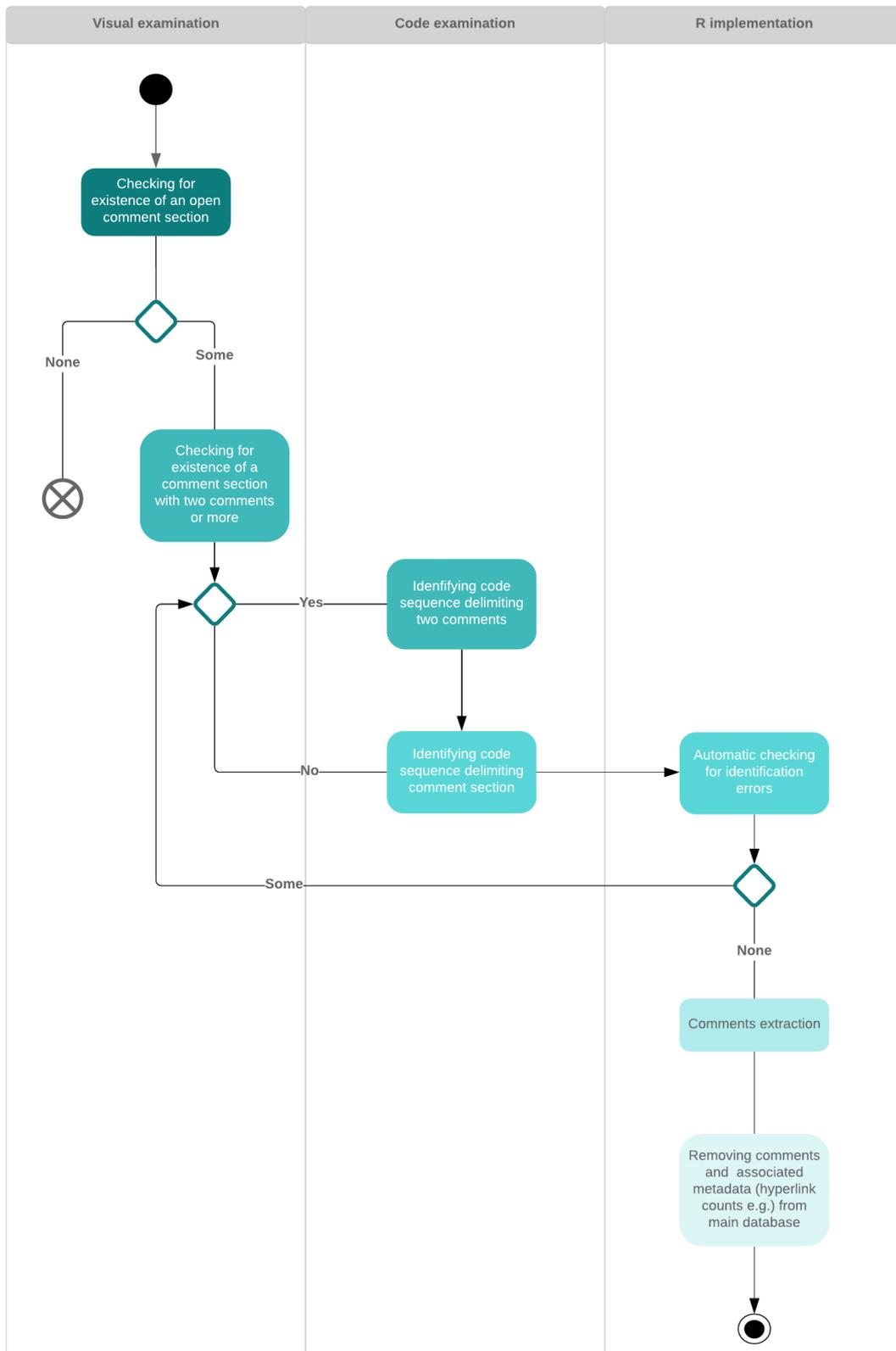

**When should we go through deletion or extraction ?**

Even following this simple procedure, deleting or extracting the commentary sections of a website can be a significant investment. To determine whether or not the operation is worth it, we simply propose to:

1. sample one's corpus of websites
2. evaluate the damage for each type of analysis planned
3. evaluate the interest of studying the commentary sections.
4. decide, according to a threshold of noise we find acceptable.

The solution we offer to this problem is of course open to improvements or amendments that would seem necessary to the specific perspective of each analyst. This is what guided our choice to publish our code on a collaborative platform. While we are aware of the amount of effort that these methodological refinements require, we are nonetheless convinced that they can prove crucial in many case studies, far beyond the specific topic of vaccination.